\documentclass{article}

\usepackage[preprint]{corl_2026} %

\usepackage{graphicx}
\usepackage{amsmath}
\usepackage{amssymb}
\usepackage{booktabs}   %
\usepackage{xcolor}     %
\usepackage{siunitx}    %
\usepackage{enumitem}   %

\usepackage{makecell} %
\usepackage{amssymb}%
\usepackage{pifont}%
\newcommand{\cmark}{\ding{51}}%
\newcommand{\xmark}{\ding{55}}%

\usepackage{float}
\usepackage{wrapfig}

\usepackage{listings}
\usepackage{tcolorbox}
\tcbuselibrary{listings,breakable}
\usepackage{capt-of} %

\title{
Slow Brain, Fast Planner: Latency-Resilient VLM-Augmented Urban Navigation
}

\author{
 Zhenghao ``Mark'' Peng$^*$\\
 Amazon FAR\\
 \And
 Honglin He\\
 UCLA\\
 \And
 Quanyi Li\\
 Independent\\
 \And
 Yukai Ma\thanks{Work done while at UCLA.}\\
 Zhejiang University\\
 \And
 Bolei Zhou\\
 UCLA\\
}

\begin{document}
\maketitle

\begin{abstract}
Learning-based planners for sidewalk navigation can generate diverse candidate trajectories in real time, yet their scoring functions often fail to select the best trajectory in challenging situations, outputting trajectories that make the mobile robot drive onto grass, toward pedestrians, or in the wrong direction, even when better candidates exist in the same set. We call this the \emph{trajectory scoring gap}: in real-world sidewalk navigation, the gap between an anchor-based planner's top choice and the best possible candidate is substantial, likely due to limited high-level scene understanding capability of the planner. Rather than replacing the planner with an end-to-end Vision-Language-Action model, we propose a VLM-Planner interface that uses a VLM to select a candidate index from the planner's proposal set and then fuse it with the planner's initial output. However, VLMs take 1--3s per query and so cannot directly drive a 5--20Hz control loop. We contribute a training-free, latency-resilient trajectory-level fusion layer that turns a stale VLM selection into real-time planner scoring via geometric similarity with exponential decay. On $\sim$2,000 challenging real-world scenarios (e.g., junctions, pedestrian encounters), VLM selection achieves 30\% ADE reduction versus the planner's best selection, while the planner remains competitive in routine situations. In simulation, Score Fusion maintains $>$80\% success rate with delays up to 5s. We demonstrate the full system on a mobile robot navigating challenging campus sidewalks with varied network latency.
\end{abstract}

\keywords{Urban Navigation, Vision Language Models}

\section{Introduction}
\label{sec:intro}
Mobile robots navigating unstructured urban environments such as sidewalks, campuses, plazas need both {reactive control} and {semantic understanding}.
Learning-based local planners~\cite{sridhar2024nomad,shah2023vint,he2025seeing} cover the reactive side well: at 5--20\,Hz they generate diverse, dynamically feasible candidate trajectories that respect the robot's kinematics and stay collision-free.
What they struggle with is {which} candidate to execute when several look feasible.
Ranking candidates for execution, \textit{i.e.} the trajectory scoring problem, remains an active research area~\cite{li2025ztrs,song2025drivecritic,shi2022mtr,kirby2026driving}.
Current approaches typically train scoring functions on geometric losses (trajectory similarity, collision proxies, smoothness) or learn from imitation data.
However, these methods inherit fundamental limitations in scene understanding: they lack the semantic reasoning required to interpret ambiguous situations, such as distinguishing among equally viable paths at junctions, understanding social norms governing pedestrian behaviors, or recognizing terrain boundaries.
We quantify this ``scoring gap'' on our $\sim$2{,}000 semantically challenging real-world scenarios as shown in Fig.~\ref{fig:normal_vs_hard}:
the planner's top choice achieves 1.64\,m ADE, while the oracle-best candidate {in the same set} achieves 0.39\,m, a gap of 1.25\,m of recoverable performance already inside the planner's output.
A VLM selector closes 30\% of this gap on the hard split while remaining competitive with the planner on routine scenarios.
The problem is not generation. It is context-dependent selection.

\begin{wrapfigure}{r}{0.55\linewidth}
    \centering
    \includegraphics[width=\linewidth]{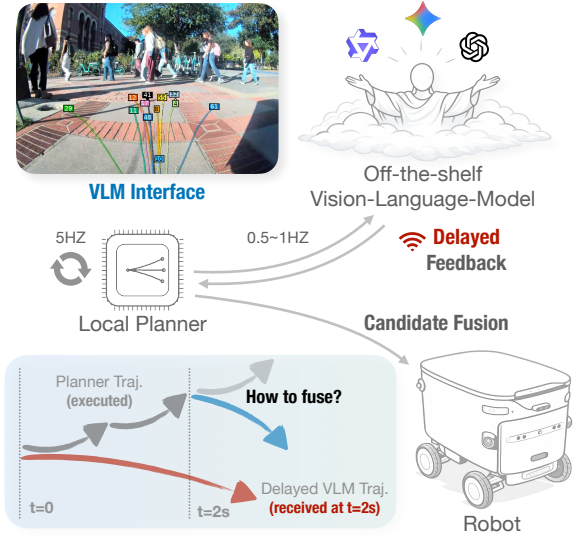}
    \vspace{-1.5em}
    \caption{\textbf{Slow Brain, Fast Planner.} A fast planner generates dynamically feasible candidates and a slow VLM selects among them. {Score Fusion} blends the stale VLM choice into the planner's real-time scoring via geometric similarity with exponential decay, enabling continuous control.}
    \label{fig:teaser}
    \vspace{-1em}
\end{wrapfigure}
Closing this scoring gap requires a source of high-level scene understanding.
Vision-Language Models (VLMs) are a good candidate: they can read scene context, understand social norms, and reason about ambiguous situations.
Recent navigation work has therefore tried to inject VLMs into the control loop, most prominently through end-to-end Vision-Language-Action models~\cite{li2025urbanvla,zitkovich2023rt,driess2023palm,brohan2023rt1} that fine-tune a VLM to output actions directly.
A more conservative line keeps the planner and uses an off-the-shelf VLM only to {select} among the planner's own candidates: PIVOT~\cite{nasiriany2024pivot} prompts a VLM with candidate annotations overlaid on the image for manipulation and navigation; VL-TGS~\cite{song2025vl} generates outdoor trajectory candidates and lets a VLM pick the best one for sidewalk-following on wheeled robots.
What both lines share, however, is a control-rate problem.
VLM inference takes 1--2\,s, orders of magnitude too slow for the 5--20\,Hz loop the robot actually runs at.
Existing systems handle this latency in three ways: (i) execute the stale VLM trajectory directly until the next response (unsafe under dynamic obstacles), (ii) decimate the VLM to low frequency and hand intermediate commands to a low-level controller~\cite{cheng2024navila}, or (iii) train a fast-slow dual system end-to-end with a learned latent interface~\cite{wei2025dualvln,wang2025internvlan1,zou2025asynchronous,chen2025fast}, which requires substantial training data and tightly couples the two stacks.
The open question is how to use a stale selection to drive a fast loop without blocking it, retraining it, or executing dangerously outdated trajectories.

Our approach takes a different path: {don't stop, but do fusion.}
Three design choices together enable latency-resilient, training-free VLM augmentation:
\textbf{(1) Planner provides dynamics; VLM provides semantics.} We constrain the VLM to picking an index $k \in \{1,\ldots,K\}$ over planner-generated candidates. The planner has already verified collision-freedom and kinematic feasibility, so any VLM choice is safe by construction.
\textbf{(2) Training-free, out-of-the-box VLMs.} We render candidates onto the camera image and prompt off-the-shelf VLMs (Gemini, GPT-5, Qwen) zero-shot, removing the need for VLA training data and letting the system absorb foundation-model improvements as they ship.
\textbf{(3) Continuous control via Score Fusion.} Stale VLM advice biases the planner's per-tick scoring rather than driving the robot directly: candidates geometrically similar to the stale selection receive a bonus that decays exponentially with age. Combined with \emph{VLM Streaming} (pipelined, fixed-cadence queries), the robot is never blocked waiting for a response.

In summary, we make the following contributions:
\begin{enumerate}
\item  We formulate VLM-augmented navigation as selection over planner candidates, enabling {training-free deployment} of off-the-shelf VLMs. On challenging scenarios, VLM selection achieves {30\% ADE reduction} versus planner argmax; on routine scenarios, the planner remains competitive, justifying our fusion approach.
\item We propose {Score Fusion} and {Probability Fusion} to handle 1--2\,s VLM latency, enabling {continuous robot motion} without blocking. In simulation, fusion maintains $>$80\% success with delays up to 5\,s; in the field, the robot operates smoothly with 1.5--3\,s network latency.
\item We demonstrate the full system on a mobile robot navigating campus sidewalks, showing that VLM augmentation reduces human interventions while preserving the planner's reactive capabilities.
\end{enumerate}

\section{Related Work}
\label{sec:related}

\noindent\textbf{Learned Local Planners.}
Modern planners generate diverse, dynamically feasible trajectories but struggle with semantic scoring.
GNM~\cite{shah2022gnm}, ViNT~\cite{shah2023vint}, and NoMaD~\cite{sridhar2024nomad} establish diffusion-based navigation with multi-hypothesis outputs.
S2E~\cite{he2025seeing} introduces anchor-based candidates, exactly the set our method consumes.
A key insight from MTR~\cite{shi2022mtr} supports our approach: their ablation shows trajectory \emph{refinement} provides marginal gains, while correct \emph{scoring} drives performance.
Rather than improving generation, we improve scoring via VLM selection.

\noindent\textbf{VLM-Augmented Trajectory Scoring and End-to-End VLAs.}
Two lines of work inject VLMs into the control loop.
The first keeps a learned planner and uses a VLM to score or select among its candidates: visual-prompt selectors that overlay numbered candidates on the image~\cite{nasiriany2024pivot,song2025vl}, scoring functions trained without imitation or from privileged information~\cite{li2025ztrs,kirby2026driving,cai2025navdp}, and VLM-derived preference, social-cost, or latent-verification signals~\cite{song2025drivecritic,hwang2025motif,song2024vlm,wu2025foresight}.
Our selection mechanism follows the visual-prompt line but is {training-free}. Our major contribution is the latency-resilient fusion layer that turns a stale selection into a usable signal at 5--20\,Hz.
The second line replaces the planner with an end-to-end VLA fine-tuned for action output~\cite{liu2025citywalker,li2025urbanvla,cheng2024navila}, which inherits 1--5\,Hz inference limits; NaVILA~\cite{cheng2024navila} in particular emits low-frequency spatial commands (``move forward 75\,cm'') for a separate locomotion policy, suitable for legged indoor settings but conservative for continuous wheeled sidewalk motion.

\noindent\textbf{Dual-System Architectures and Latency Handling.}
A growing set of designs handles VLM/VLA latency via {action-level} asynchrony, where a slow stack emits action chunks or latent commands and a fast head stitches them together.
Slow-fast architectures pair semantic reasoning at 1--2\,Hz with reactive control at 20--30\,Hz through learned latent plans or jointly-trained fast experts~\cite{chen2025fast,han2024dual,figure2025helix,wei2025dualvln,wang2025internvlan1,zou2025asynchronous,black2025pi05,nvidia2025groot}.
A second cluster handles the action-chunk boundary explicitly via polynomial seam smoothing or forward-rolled-state queries~\cite{zhao2025vla,tang2025vlash}, while StreamVLN~\cite{wei2025streamvln} attacks VLM-internal latency through slow-fast context modeling and KV-cache reuse.
FOREWARN~\cite{wu2025foresight} is closest in spirit, using a learned latent world model to let the VLM verify rollouts.
All of these require either training a shared interface or committing to the seam of an old action chunk.
Our fusion operates at the {trajectory level}: every control tick the planner regenerates a candidate set on fresh observations, and fusion picks the current candidate whose geometry best matches the stale VLM choice with exponential staleness decay.
The mechanism is heuristic and training-free, requiring no joint optimization with planner or VLM.

Tab.~\ref{tab:related_work_comparison} positions our approach against these prior works: among methods that retain a learned planner, ours is the only one that is simultaneously training-free, continuous-control, and latency-resilient via trajectory-level fusion (rather than action-level chunk stitching).

\section{Method}
\label{sec:method}

We propose a hybrid architecture that combines a fast local planner and a slow Vision-Language Model (VLM).
The planner generates dynamically feasible candidate trajectories at high frequency and the VLM provides semantic judgment asynchronously with 1--2\,s latency.
The key challenge is bridging this temporal mismatch: how can stale VLM advice improve real-time trajectory selection?

\subsection{System Overview}
\label{sec:system_overview}

Our system operates as two loosely-coupled loops (Fig.~\ref{fig:teaser}):

\noindent \textbf{Fast Loop (Planner).}
At each control tick (5--20\,Hz), the planner generates $K$ candidate trajectories $\{\tau_1, \ldots, \tau_K\}$, each a sequence of $(x, y)$ waypoints in the robot frame.
The planner is trained to emit candidates that are kinematically realizable and locally obstacle-avoiding, and assigns each candidate a confidence score from its internal objective.
We denote the planner scores as $S_1(\tau_i)$.
This means the planner already absorbs robot dynamics and short-horizon avoidance; the VLM's role downstream is to break ties on semantically ambiguous picks among an already-feasible set.

\noindent \textbf{Slow Loop (VLM).}
A VLM asynchronously receives a visual query: the current camera image with $K$ candidate trajectories overlaid as colored polylines.
The VLM outputs a selected index $k^\star \in \{1, \ldots, K\}$, representing its judgment of which trajectory best satisfies the navigation goal and social norms.
This call takes $\Delta t \approx 1$--$2$\,s; by the time the response arrives, the robot has moved and the planner has generated new candidates.

\noindent \textbf{Fusion.}
Rather than directly executing the (now stale) VLM-selected trajectory, we use its \emph{semantic intent} to bias the current planner candidates.
The VLM's choice encodes high-level preferences (``stay on sidewalk,'' ``yield to pedestrian'') that remain valid even as the specific trajectory becomes outdated.
We propose \emph{latency-resilient fusion} mechanisms that combine geometric similarity with exponential staleness decay.

\begin{figure}[!t]
\centering
\begin{minipage}[b]{0.57\linewidth}
\centering
\captionof{table}{Comparison with related VLM-augmented robot control approaches.}
\label{tab:related_work_comparison}
\setlength{\tabcolsep}{3pt}
\renewcommand{\arraystretch}{1.15}
\resizebox{\linewidth}{!}{%
\begin{tabular}{lccccc}
\toprule
Method & \makecell{Training\\Free} & \makecell{Continuous\\Control} & \makecell{Latency\\Handling} & \makecell{Fusion\\Level} & \makecell{Planner\\Fallback} \\
\midrule
E2E VLAs~\cite{zitkovich2023rt,black2024pi0} & \xmark & \xmark & \xmark & -- & \xmark \\
NaVILA~\cite{cheng2024navila} & \xmark & \xmark & Low-freq cmd & Action & \cmark \\
PIVOT~\cite{nasiriany2024pivot} & \cmark & \xmark & \xmark & Traj. & \xmark \\
VL-TGS~\cite{song2025vl} & \cmark & \xmark & \xmark & Traj. & \xmark \\
VLM-Social-Nav~\cite{song2024vlm} & \cmark & \xmark & \xmark & Score & \xmark \\
FOREWARN~\cite{wu2025foresight} & \xmark & -- & -- & Latent traj. & \cmark \\
VLA-RAIL~\cite{zhao2025vla} & \xmark & \cmark & Chunk stitch & Action & \xmark \\
VLASH~\cite{tang2025vlash} & \xmark & \cmark & Future-state & Action & \xmark \\
DuoCore-FS~\cite{zou2025asynchronous} & \xmark & \cmark & Async latent & Action & \xmark \\
$\pi_{0.5}$, GR00T~\cite{black2025pi05,nvidia2025groot} & \xmark & \cmark & Async & Action & \xmark \\
\midrule
\textbf{Ours} & \cmark & \cmark & Score Fusion & \textbf{Traj.} & \cmark \\
\bottomrule
\end{tabular}%
}
\end{minipage}\hfill
\begin{minipage}[t]{0.4\linewidth}
\centering
\includegraphics[width=\columnwidth]{figs/main_fig.pdf}
\caption{VLM selection performance. The planner is better on normal and the VLM outperforms on hard.}
\label{fig:normal_vs_hard}
\end{minipage}
\end{figure}

\subsection{Visual Trajectory Selection}
\label{sec:visual_prompting}

We render candidate trajectories onto the current camera image as numbered, colored annotations and let an off-the-shelf VLM pick an index.

\noindent \textbf{Visual Overlay.}
We project each candidate trajectory $\tau_i$ onto the camera image using the known camera extrinsics and intrinsics.
Each trajectory is rendered as a colored polyline with its index labeled at the endpoint.
The goal direction is optionally marked with an arrow.
This visual representation lets the VLM reason directly in pixel space: it can see where each trajectory leads relative to sidewalk boundaries, pedestrians, and obstacles.

\vspace{0.25em}
\noindent \textbf{Structured Output.}
The VLM outputs a JSON response specifying its selected index and reasoning (see Fig.~\ref{fig:qualitative}).
The VLM can also select action ``stop'' as a safety guardrail.
If the VLM cannot parse a valid selection or times out, the system falls back to the planner's argmax: $k = \arg\max_i S_1(\tau_i)$.

\noindent \textbf{Training-Free Deployment.}
Critically, we use off-the-shelf VLMs (Gemini, GPT-5, Qwen) without any fine-tuning.
The visual prompting interface transforms trajectory selection into a visual reasoning task that general-purpose VLMs can solve zero-shot.
This eliminates the need for VLA training data, domain adaptation, or specialized model architectures.

\subsection{Latency-Resilient Fusion}
\label{sec:fusion}

The central question is how to use a delayed VLM selection as a usable signal for real-time control: rather than discarding it or executing the outdated trajectory itself, we fuse its \emph{semantic intent} into the planner's per-tick scoring over freshly generated candidates.
Below we describe the shared foundation (geometric similarity, staleness decay) and two fusion variants.

\begin{figure}[!t]
\centering
\includegraphics[width=\linewidth]{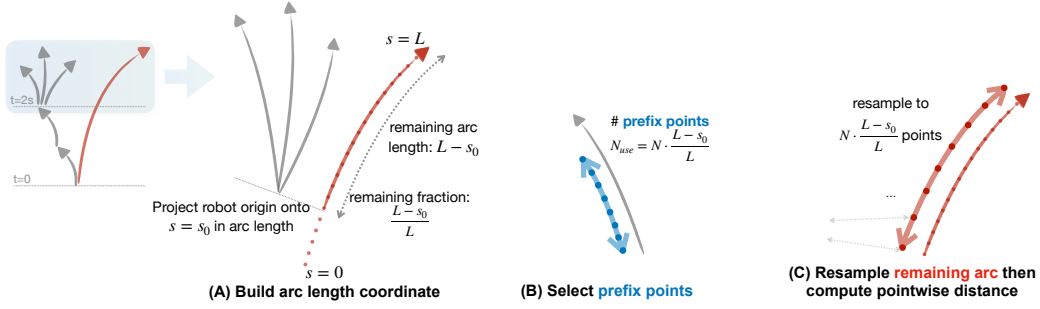}
\vspace{-1.5em}
\captionof{figure}{\textbf{Horizon-aware similarity for delayed VLM guidance.} The stale VLM trajectory is motion-compensated into the current frame; arclength progress $s_0$ defines the remaining horizon over which similarity is computed (Sec.~\ref{sec:fusion}).}
\label{fig:method}
\vspace{-1em}
\end{figure}

\noindent \textbf{Geometric Similarity.}
Let $\tau_{\text{vlm}}$ denote the trajectory the VLM selected (in the robot frame at query time).
We motion-compensate it into the \emph{current} robot frame using odometry, then compute a horizon-aware similarity (Fig.~\ref{fig:method}):
\begin{equation}
    \text{Sim}(\tau_i, \tau_{\text{vlm}}) =
    -\frac{1}{d_{\text{scale}}}\cdot \frac{1}{N_{\text{use}}}\sum_{n=1}^{N_{\text{use}}}
    \left\lVert \tau_i[n] - \hat{\tau}_{\text{vlm}}[n] \right\rVert_2
    ,
    \label{eq:similarity}
\end{equation}
where $\hat{\tau}_{\text{vlm}}$ is the \emph{remaining} segment of the delayed VLM trajectory after progress alignment.
Specifically, we project the robot origin onto the motion-compensated polyline to obtain arclength progress $s_0$ and remaining fraction $f=(L-s_0)/L$, set $N_{\text{use}}=\mathrm{round}(N\cdot f)$, and compare only the overlapping horizon.
Higher similarity means the candidate follows the VLM's intent without forcing a match on the future tail.

\noindent \textbf{Staleness Decay.}
As VLM advice ages, its relevance decreases. We apply exponential decay:
$
    w(\Delta t) = \exp\left(-\frac{\Delta t}{\tau_{\text{decay}}}\right)
$
where $\tau_{\text{decay}}$ is a time constant (typically 3--5\,s).
Fresh advice receives full weight; advice older than $\sim 2\tau_{\text{decay}}$ is effectively ignored.

\noindent \textbf{Score Fusion.}
The simplest approach adds a weighted similarity term to the planner's scores:
\begin{equation}
    S_{\text{fused}}(\tau_i) = S_1(\tau_i) + \lambda \cdot w(\Delta t) \cdot \text{Sim}(\tau_i, \tau_{\text{vlm}}),
    \label{eq:score_fusion}
\end{equation}
where $\lambda$ controls VLM influence strength.
The robot executes $k = \arg\max_i S_{\text{fused}}(\tau_i)$.
This formulation gracefully degrades: without VLM advice, it reduces to planner argmax; stale advice decays naturally.

\noindent \textbf{Probability Fusion.}
Score Fusion's unbounded similarity term can dominate when $\lambda$ is large.
\emph{Probability Fusion} provides a more controlled alternative by mixing distributions:
\begin{equation}
    p_{\text{fused}}(\tau_i) = (1 - \alpha) \cdot p_{\text{planner}}(\tau_i) + \alpha \cdot p_{\text{vlm}}(\tau_i),
    \label{eq:prob_fusion}
\end{equation}
where $p_{\text{planner}}$ and $p_{\text{vlm}}$ are softmax distributions over planner scores and VLM similarities respectively, and $\alpha = \frac{\lambda}{\lambda + 1} \cdot w(\Delta t)$ is a staleness-dependent mixing coefficient.
Even with $\lambda \to \infty$, the coefficient $\alpha \leq 1$, bounding the influence of VLM on planner.

\noindent \textbf{VLM Streaming.}
We submit queries at a fixed cadence (1\,Hz by default) without waiting for previous responses, keep multiple requests in flight, and on each tick fuse with the {newest-by-timestamp} response that has arrived (out-of-order arrivals are common over cellular).
The robot is never blocked: the fusion mechanism always has {some} semantic guidance available.
This differs from prior dual-system designs~\cite{cheng2024navila,wei2025dualvln,wang2025internvlan1} that handle latency by decimating the slow loop and committing to a specific low-level policy or action vocabulary. Our streaming-plus-fusion path keeps the slow loop at its native rate and uses the planner's regenerated candidates as the live interface, so the VLM and the planner remain independently swappable.

\begin{figure}[!t]
\centering
\includegraphics[width=\linewidth]{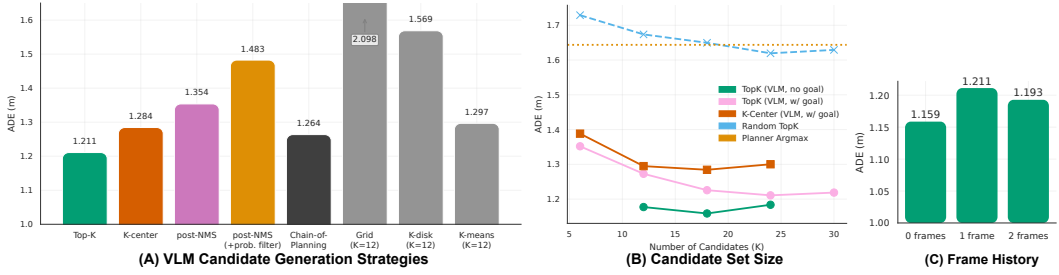}
\vspace{-1.5em}
\captionof{figure}{\textbf{Ablations on VLM trajectory selection (1,412 hard scenarios).}
\textbf{(A)} Candidate generation: score-based Top-K vs.\ geometric diversification.
\textbf{(B)} Candidate set size $K$.
\textbf{(C)} Prompt content: goal, scores, and history frames.}
\label{fig:trajectory_selection_main}
\vspace{-1em}
\end{figure}

\begin{table*}[t]
\centering
\caption{VLM trajectory selection quality and cost across model families. All models use Top-K=18 candidates. Latency is median end-to-end inference time in seconds.}
\label{tab:planning_quality}
\setlength{\tabcolsep}{3.5pt}
\renewcommand{\arraystretch}{1.15}
\resizebox{\textwidth}{!}{%
\begin{tabular}{l
c c c
c c c
c c c c
}
\toprule
& \multicolumn{3}{c}{\textbf{Trajectory Quality}} & \multicolumn{3}{c}{\textbf{Goal Alignment}} & \multicolumn{4}{c}{\textbf{Cost / Latency}} \\
\cmidrule(lr){2-4} \cmidrule(lr){5-7} \cmidrule(lr){8-11}
Model
& {ADE$\downarrow$} & {ADE@1s$\downarrow$} & {ADE@2s$\downarrow$}
& {AngDiff$\downarrow$} & {Progress$\uparrow$} & {MAOE$\downarrow$}
& {In Tok} & {Out Tok} & {Total Tok} & {Latency (s)$\downarrow$} \\
\midrule
Planner Argmax           & 1.64 & 0.64 & 1.06 & 16.08 & 4.88 & 16.38 & {--} & {--} & {--} & {--} \\
\midrule
Gemini 3 Flash           & \textbf{1.16} & \textbf{0.39} & \textbf{0.66} & 16.76 & 6.16 & \textbf{15.70} & 2744 & 62 & 5957 & 8.1 \\
Gemini 2.5 Flash         & 1.20 & 0.41 & 0.69 & 17.04 & \textbf{6.32} & 16.14 & 1902 & 82 & 4134 & 13.2 \\
Gemini 2.5 Flash Lite    & 1.21 & 0.40 & 0.68 & 17.43 & 6.20 & 16.50 & 1902 & 85 & 1994 & \textbf{1.7} \\
Gemini Robotics ER 1.5   & 1.21 & 0.41 & 0.69 & 16.78 & 6.06 & 15.86 & 1902 & 73 & 3372 & 14.4 \\
\midrule
GPT-5 Mini               & 1.47 & 0.48 & 0.82 & 17.53 & 5.05 & 18.00 & 2181 & {--} & 3196 & 16.6 \\
GPT-5 Nano               & 1.77 & 0.54 & 0.98 & 17.53 & 4.43 & 17.33 & 2334 & {--} & 5053 & 22.0 \\
\midrule
Qwen2.5-VL-72B           & 1.20 & 0.40 & 0.68 & 17.11 & 6.01 & 16.18 & 2250 & 43 & 2293 & 11.8 \\
Qwen2.5-VL-32B           & 1.81 & 0.55 & 0.99 & 17.20 & 4.12 & 17.41 & 2250 & 52 & 2302 & 13.1 \\
Qwen2.5-VL-7B            & 2.27 & 0.69 & 1.25 & \textbf{16.46} & 3.33 & 19.45 & 2250 & 48 & 2298 & 2.7 \\
\bottomrule
\end{tabular}
}
\vspace{-1em}
\end{table*}

\section{Experiments}
\label{sec:experiments}

The evaluation has three components:
(1)~\emph{offline trajectory selection} on real-world navigation logs;
(2)~\emph{closed-loop simulation under VLM latency} with a controlled corrupted planner, isolating the effect of latency and fusion policy;
(3)~a \emph{real-world deployment} on campus sidewalks under realistic cellular latency.

\subsection{Offline Trajectory Selection}
\label{sec:exp_offline}

We test whether an off-the-shelf VLM-as-selector helps on our real-world sidewalk dataset, which sets an upper bound on what fusion can preserve under latency.

\noindent \textbf{Dataset.}
We use two pools of logs from wheeled robots on urban sidewalks. The \emph{normal} pool ($\sim$3{,}000 snapshots) is sampled routes on relatively unchallenging streets, where the planner's argmax is typically already a sensible action. The \emph{hard} pool ($\sim$2{,}000 snapshots) is collected on a university campus and curated to over-represent semantically challenging situations: junctions, pedestrian and cyclist encounters, terrain boundaries (grass, planters, curbs), obstacles, and ambiguous forks. Each snapshot ships an RGB frame, the 64 anchor-based candidate trajectories with planner scores, and the human teleoperator's executed path as ground truth.

\noindent \textbf{Local Planner.}
We use S2E~\cite{he2025seeing}, an anchor-based model that emits 64 multi-hypothesis $(x,y)$ trajectories per frame with planner scores. Full details are in the appendix.

\noindent \textbf{Baselines and Metric.}
We compare \textit{Planner Argmax} (the planner's own top-1) and \textit{Oracle} (the min-ADE candidate among the planner's 64 raw candidates, an upper bound for any selector that consumes this candidate set). We report \emph{Average Displacement Error} (ADE), the mean Euclidean distance between the selected trajectory's waypoints and the human teleoperator's path over the candidate horizon, as our primary metric.

\noindent \textbf{Main Results: VLM Excels in Hard Scenarios.}
Fig.~\ref{fig:normal_vs_hard} reports ADE on both splits.
On the {hard} split (Fig.~\ref{fig:normal_vs_hard}B), the VLM selector (Gemini 3 Flash) achieves {1.16\,m ADE}, reducing error by 30\% compared to the planner argmax (1.64\,m) and approaching the oracle lower bound (0.39\,m); the planner's scoring function struggles with semantic ambiguity, while the VLM's visual reasoning handles these situations well.
On the {normal} split (Fig.~\ref{fig:normal_vs_hard}A), the VLM actually {underperforms} the planner: in routine situations the planner's learned scoring is already well-calibrated, and VLM selection introduces unnecessary variance. The oracle confirms that better candidates exist on this split too, but the planner already selects near-optimally.

\begin{figure}[!t]
\centering
\begin{minipage}[t]{0.5\linewidth}
\centering
\includegraphics[width=\linewidth]{figs/quali.pdf}
\vspace{-1.5em}
\captionof{figure}{
\textbf{(A)} At a junction, the VLM picks trajectory~14 along the sidewalk curve.
\textbf{(B)} On a routine straight, the VLM applies near-identical reasoning but again picks 14, while the planner's argmax 7 was better.}
\label{fig:qualitative}
\end{minipage}\hfill
\begin{minipage}[t]{0.47\linewidth}
\centering
\includegraphics[width=\linewidth]{figs/closed_loop_sim.pdf}
\vspace{-1.5em}
\captionof{figure}{\textbf{Closed-loop success rate vs.\ VLM delay}. Score Fusion holds above 80\% out to 5\,s and Probability Fusion stays within a few points of it ($\sim$78\% at 5\,s), while VLM Hold collapses past 2\,s and VLM Stream past 3\,s.}
\label{fig:closed_loop_sim}
\end{minipage}
\vspace{-1em}
\end{figure}

This finding is consistent with observations in autonomous driving, where VLM-based systems excel at handling edge cases but may overthink routine situations~\cite{tian2024drivevlm,song2025drivecritic}.
The pattern also shows up qualitatively, as shown in Fig.~\ref{fig:qualitative}: in the success case~(A), the VLM correctly identifies the sidewalk curve at a junction, improving FDE by 3.2\,m; in the failure case~(B), the VLM applies similar reasoning to a routine straight segment but selects a suboptimal trajectory, degrading FDE by 4.5\,m.
This motivates our \emph{fusion} approach: rather than always trusting the VLM, we blend its advice with the planner's judgment so that the VLM's semantic understanding dominates in hard scenarios while the planner's confidence remains the anchor in normal ones.

\noindent \textbf{Score-Based Filtering Outperforms Geometric Diversification.}
As shown in Fig.~\ref{fig:trajectory_selection_main}(A), we compared six strategies for selecting $K$ candidates from the original 64.
$\bullet$ \textit{Top-K} (our default) keeps the $K$ highest-scored candidates; 
$\bullet$ \textit{K-center} runs greedy farthest-point sampling on trajectory endpoints to maximize geometric diversity; 
$\bullet$ \textit{K-means} clusters endpoints and returns medoids; 
$\bullet$ \textit{Post-NMS} suppresses near-duplicate candidates by trajectory similarity; 
$\bullet$ \textit{Grid} ignores the planner entirely and uses static prototype trajectories defined by an angle/distance grid with linear interpolation; and $\bullet$ \textit{Chain-of-Planning} applies a two-stage hierarchical selection on top of the others.
Score-based Top-K filtering is the best of these; K-center and K-means achieve similar performance, while static grid prototypes perform poorly.
The takeaway is that {planner confidence provides valuable signal for candidate pruning}: the top-scored candidates are more likely to contain good options than geometrically diverse samples are.

\noindent \textbf{Scaling the candidate set size $K$.}
As shown in Fig.~\ref{fig:trajectory_selection_main}(B), VLM selection improves as $K$ grows from 6 to 18 and then plateaus. Beyond $K\approx24$ additional candidates make the visual prompt more complex without adding value.

\noindent \textbf{Prompt Design.}
All prompt-design ablations use the hard subset (1,412 snapshots) where selection is most challenging, as shown in Fig.~\ref{fig:trajectory_selection_main}(C).
First, adding history frames (1--2 previous camera images) {degrades} performance slightly, as the planner candidates already encode short-horizon feasibility and the VLM's role is semantic understanding, which is fully captured in the current frame.
Second, hiding goal information and planner scores {improves} selection: when scores are visible, the VLM tends to defer to the planner's ranking rather than apply its own judgment.
Our best configuration hides both, letting the VLM reason from visual evidence alone.

\noindent \textbf{Model Comparison.}
Tab.~\ref{tab:planning_quality} compares VLM trajectory selection across model families on the hard scenarios.
Gemini 3 Flash achieves the best overall ADE, closely followed by Gemini 2.5 Flash variants and Qwen2.5-VL-72B.
GPT-5 models show weaker performance, and smaller Qwen models degrade further.
For real-time deployment, \textbf{Gemini 2.5 Flash Lite} offers the best latency--quality tradeoff: 1.7\,s median latency with competitive ADE.

\subsection{Closed-Loop Simulation Under VLM Latency}
\label{sec:exp_simulation}

We study whether VLMs are still helpful under 1--3\,s of latency in closed loop, and whether {fusion} preserves the headroom that direct execution of stale VLM trajectories does not.

\noindent \textbf{Setup.}
We develop a lightweight simulator that uses human-teleoperation trajectories as the reference and a fixed set of $K{=}12$ trajectories as the ``planner's'' candidate set.
The ``VLM'' is a \emph{delayed oracle}: at query time it picks the min-ADE candidate and returns $\Delta t \in \{0, 0.5, \ldots, 5.0\}$\,s later, simulating the latency.
The planner's score is corrupted so Local-only does not trivially pick the best candidate.
We compare \emph{Planner Oracle} (uncorrupted argmax), \emph{Local-only} (corrupted argmax), \emph{VLM Hold/Stream}, \emph{VLM Match}, and \emph{Score/Probability Fusion} (Eq.~\ref{eq:score_fusion}, \ref{eq:prob_fusion}); full details in App.~\ref{sec:appendix_closed_loop}.
The metric is success rate---the fraction of episodes reaching the reference endpoint.

\noindent \textbf{Main result.}
Fig.~\ref{fig:closed_loop_sim} plots success rate as a function of $\Delta t$.
Score Fusion holds above 80\% out to 5\,s of VLM latency and Probability Fusion remains close to that level (around 78\% at 5\,s), while VLM Hold collapses past 2\,s and is near zero by 4\,s; streaming buys VLM Stream a few additional seconds but it still falls below 20\% success at 5\,s.
Fusion outperforms both stale-execution policies because the planner regenerates candidates each tick on fresh observations and the geometric similarity term picks the current candidate whose geometry best matches the stale VLM's semantic intent, instead of committing to a polyline the world has already moved past.

\noindent \textbf{Probability Fusion is more stable.}
With streaming, Score Fusion can become unstable at large $\lambda$ because its similarity term is unbounded; Probability Fusion bounds VLM influence via $\alpha \leq 1$ and is markedly less sensitive to hyperparameter choice.
We therefore recommend Probability Fusion with streaming for deployment, which is what our real-robot setup uses in Sec.~\ref{sec:exp_real_robot}.

\subsection{Real-World Deployment}
\label{sec:exp_real_robot}

\begin{figure}[!t]
\centering
\includegraphics[width=\linewidth]{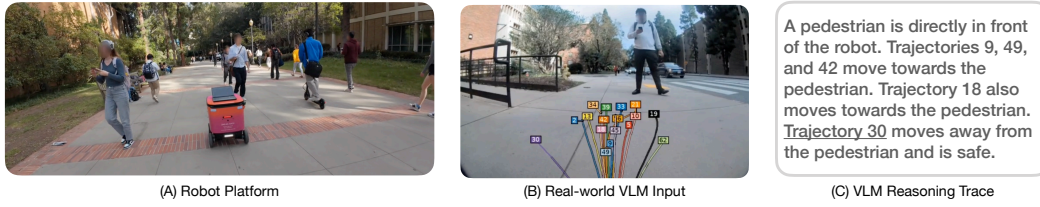}
\vspace{-2em}
\caption{\textbf{Real-world deployment.}
\textbf{(A)~Robot platform.} We use a four-wheeled delivery robot on a campus sidewalk shared with pedestrians.
\textbf{(B)~Real-world VLM input.} The actual overlay the VLM sees: the front-camera frame with the trajectory overlay and a pedestrian is directly ahead.
\textbf{(C)~VLM reasoning trace.} The natural-language \texttt{reason} field by VLM.}
\label{fig:real_world_robot}
\vspace{-1em}
\end{figure}

We validate the full system on a four-wheeled delivery robot navigating urban sidewalks (Fig.~\ref{fig:real_world_robot}). Closed-loop trajectories, VLM responses, and the takeover events that drive the safety metrics in Tab.~\ref{tab:real_robot_metrics} are shown in the supplementary video.

\noindent \textbf{Robot and Control.}
We use a four-wheeled robot equipped with an RGB camera and wheel odometry.
At each control tick, the local planner outputs a short-horizon waypoint polyline, and a pure pursuit controller converts waypoints into linear and angular velocity commands $(v,\omega)$.
Odometry provides real-time state estimation for closed-loop tracking and goal reference updates.
All policies run at 5\,Hz.
The local planner executes on a laptop with an NVIDIA RTX~3080 mobile GPU.
VLM queries are sent over a 4G cellular connection to a cloud API (Gemini 2.5 Flash Lite), with observed round-trip latency typically 1.5--3.0\,s depending on network conditions.

\noindent \textbf{Evaluation Protocol.}
We evaluate on 5 real-world routes covering diverse urban sidewalk environments.
Each policy is tested across multiple runs.
A human operator monitors the robot and intervenes when the robot is about to drive off the sidewalk, contact an obstacle, or otherwise enter an unsafe state.
We report metrics in three categories in Tab.~\ref{tab:real_robot_metrics}:
1) \textit{Safety metrics} measure human intervention frequency.
``Takeovers per 100\,m'' and ``per minute'' normalize interventions by distance and time respectively.
``Longest autonomous segment'' measures the maximum distance traveled without intervention.
2)
\textit{Trajectory consistency metrics} measure execution stability.
``Temporal consistency'' (TempCons) is the motion-compensated ADE between consecutive waypoint outputs; lower values indicate smoother trajectories.
``Exec--Argmax deviation'' measures how far the executed trajectory deviates from the planner's current argmax, indicating the VLM's influence on execution.
3)
\textit{VLM agreement metrics} characterize VLM--planner alignment.
``VLM--Argmax deviation'' measures the geometric deviation between VLM-selected and planner-selected trajectories at query time.
``Agreement rate'' is the percentage where VLM and planner select the same trajectory index.

\begin{table*}[!t]
\centering
\caption{Real-robot closed-loop metrics. All deviation metrics report 90th percentile values.}
\label{tab:real_robot_metrics}
\setlength{\tabcolsep}{4.5pt}
\renewcommand{\arraystretch}{1.15}
\resizebox{\textwidth}{!}{%
\begin{tabular}{lccc|cc|ccc}
\toprule
& \multicolumn{3}{c|}{\textbf{Safety}} & \multicolumn{2}{c|}{\textbf{Trajectory Consistency}} & \multicolumn{3}{c}{\textbf{VLM--Planner Alignment}} \\
\cmidrule(lr){2-4} \cmidrule(lr){5-6} \cmidrule(lr){7-9}
Method &
\makecell[c]{Takeovers\\per 100m$\downarrow$} &
\makecell[c]{Takeovers\\per min$\downarrow$} &
\makecell[c]{Longest\\Seg (m)$\uparrow$} &
\makecell[c]{TempCons\\(m)$\downarrow$} &
\makecell[c]{Exec--Argmax\\Dev (m)$\downarrow$} &
\makecell[c]{Match--Argmax\\Dev (m)} &
\makecell[c]{VLM--Argmax\\Dev (m)} &
\makecell[c]{Agreement\\(\%)} \\
\midrule
Local-only & 3.49 & 0.99 & 88.4 & 0.57 & 0.00 & -- & -- & -- \\
VLM Hold & 8.15 & 1.19 & 27.0 & 0.07 & 3.99 & -- & 2.90 & 10.7 \\
VLM Stream & 4.40 & 1.15 & 71.9 & 0.48 & 3.36 & -- & 3.35 & 9.2 \\
VLM Match (Hold) & 1.74 & 0.46 & 88.8 & 0.47 & 3.42 & 4.02 & 3.94 & 5.6 \\
VLM Match (Stream) & 2.18 & 0.59 & 87.1 & 0.47 & 3.16 & 3.64 & 3.53 & 8.5 \\
Score Fusion (Stream) & 1.31 & 0.37 & 95.2 & 0.36 & 1.77 & 3.51 & 3.61 & 7.1 \\
\textbf{Prob Fusion (Stream)} & \textbf{0.87} & \textbf{0.24} & \textbf{111.7} & 0.45 & 1.56 & 3.86 & 3.40 & 9.1 \\
\bottomrule
\end{tabular}
}
\vspace{-1.5em}
\end{table*}

\noindent \textbf{Results.}
1) \textit{Naive VLM execution is dangerous.}
VLM Hold produces the highest intervention rate (8.15 takeovers per 100\,m) and the shortest autonomous segment (27.0\,m).
The robot executes trajectories selected seconds ago, leading to collisions or off-road excursions.
2) \textit{Fusion enables safe continuous control.}
Probability Fusion with Streaming reduces interventions to {0.87 per 100\,m}, a 75\% reduction from Local-only (3.49) and 89\% from VLM Hold (8.15).
It also achieves the longest autonomous segment (111.7\,m) and bounded deviation from the planner's argmax (1.56\,m).
3) \textit{Streaming alone is insufficient.}
VLM Stream (without fusion) still suffers elevated interventions (4.40/100\,m) because it directly executes stale trajectories.
Fusion is necessary to translate VLM advice into safe real-time control.

Human observers noted that the robot's behavior appeared ``intentional'' rather than ``confused,'' a qualitative improvement over the planner-only baseline. Full closed-loop traces of real-world experiments are shown in the supplementary video.

\section{Conclusion}
\label{sec:conclusion}

We have presented a latency-resilient approach to VLM-augmented navigation that enables continuous robot control despite 1--2\,s VLM inference delays.
Our key insight is that a fast planner and a slow VLM provide complementary capabilities that can be fused rather than forced into a single system.
Off-the-shelf VLMs excel at trajectory selection in semantically challenging scenarios (30\% ADE reduction), while learned planners remain competitive in routine situations---motivating a fusion approach rather than VLM-only control.
Score and Probability Fusion enable continuous control under latency.
Real-world deployment with Probability Fusion and VLM Streaming substantially reduces human interventions compared to both planner-only and naive VLM execution.

\paragraph{Limitations and Future Work}
Our approach inherits the planner's candidate set; if no good candidate exists, the VLM cannot help.
We find VLM selection does not universally outperform the planner: in routine scenarios, the planner's learned scoring often suffices, and VLM queries consume computational resources without benefit.
Our closed-loop simulator also models the VLM as a delayed oracle, which cannot exhibit real-world scene drift.
Next steps include adaptive querying that invokes the VLM only when the planner is uncertain, 
studying the interaction between VLM and planner beyond trajectory selection, and testing the whole system in more realistic simulator.

\bibliography{references}

\clearpage

\clearpage
\appendix

\newtcblisting{promptbox}{
  breakable,
  colback=black!2,
  colframe=black!20,
  arc=10pt,
  boxrule=0.8pt,
  left=10pt,
  right=10pt,
  top=5pt,
  bottom=5pt,
  listing only,
  listing engine=listings,
  listing options={
    basicstyle=\ttfamily\scriptsize,
    breaklines=true,
    showstringspaces=false,
    escapeinside={(*@}{@*)}
  }
}

\section{VLM Interface for Trajectory Selection}
\label{sec:appendix_vlm_interface}

\subsection{Interface Overview}
\label{sec:appendix_vlm_interface_overview}

Our VLM does \emph{trajectory selection} rather than low-level control: at each step, a fast local planner proposes a discrete set of short-horizon candidate trajectories (4\,s horizon), and the VLM returns either (i) the index of one candidate to execute next or (ii) a stop decision when none appear safe. This constrains VLM outputs to dynamically feasible motions and enables safety fallbacks.

\subsection{Local Planner: Anchor-Based Candidate Generation}
\label{sec:appendix_planner}

We adopt S2E~\cite{he2025seeing} as our local planner.
Unlike diffusion-based navigation models (e.g., NoMaD~\cite{sridhar2024nomad}), S2E uses \emph{anchor-guided distribution matching} to produce a structured set of candidate trajectories.

\noindent \textbf{Anchor set.}
The model defines $M=64$ anchor points obtained via k-means clustering over trajectory endpoints in the training data.
Each anchor represents a prototypical behavioral mode (e.g., go straight, veer left, slow down, sharp turn).
These anchors are fixed after training and serve as queries in a cross-attention decoder.

\noindent \textbf{Architecture and outputs.}
Given the current RGB observation (past $k=5$ frames) and a goal coordinate, an EfficientNet encoder and a Transformer encoder produce scene context embeddings.
A Transformer decoder then cross-attends from the $M$ anchor queries to these context embeddings, producing per-anchor features.
Three lightweight heads decode each anchor feature into:
\begin{enumerate}[leftmargin=*,nosep]
    \item a \textbf{score} $q_m \in [0,1]$ (softmax-normalized), representing the model's confidence that anchor $m$ is the best behavioral mode for the current situation;
    \item a \textbf{regression trajectory} $\tau_m$: a sequence of $(x,y)$ waypoints (normalized offsets from the anchor), forming a 4\,s, 20-waypoint polyline in the robot frame;
    \item a \textbf{velocity scale} $v_m$, converting the normalized trajectory into metric coordinates.
\end{enumerate}
The result is 64 candidate trajectories, each with an associated planner score.
In our pipeline, we select the top-$K$ candidates by score (default $K{=}18$) for presentation to the VLM (see ablation in the main paper, Fig.~\ref{fig:trajectory_selection_main}(B)).

\noindent \textbf{Why anchor-based design matters for our method.}
The anchor-based architecture is particularly well-suited to our VLM-augmented pipeline for two reasons.
First, the \emph{scores} provide a natural baseline ranking that our fusion mechanism can augment: Score Fusion (Eq.~\ref{eq:score_fusion}) adds a VLM similarity bonus to these planner scores, so the planner's learned confidence is preserved and the VLM only overrides it when semantic judgment disagrees.
Second, because the anchors are \emph{fixed} across time steps, consecutive frames produce geometrically consistent candidate sets, which makes the horizon-aware similarity computation for stale VLM advice (Sec.~\ref{sec:fusion}) more stable than it would be with a diffusion-based planner whose candidates vary stochastically from frame to frame.

\subsection{Candidate Visualization (Overlay Design)}
\label{sec:appendix_overlay_design}

\noindent \textbf{What is rendered.}
We render an overlay on the front camera image with:
\begin{itemize}[leftmargin=*,nosep]
    \item candidate trajectories as colored polylines (or optionally as swept-footprint corridors);
    \item a small dot at each candidate endpoint;
    \item an integer index label near each endpoint (the label text is the authoritative ID);
    \item an optional goal cue (magenta \texttt{GOAL} marker and/or a ``hanging'' arrow).
\end{itemize}

\noindent \textbf{Projection and geometry.}
Candidates are defined in the robot body frame $(x_\text{forward}, y_\text{left})$ on the ground plane and projected into the camera image using a lightweight fisheye projection. The overlay legend reminds the VLM that fisheye distortion is normal near image borders.

\noindent \textbf{Label-to-line disambiguation.}
To reduce index confusion when trajectories overlap, each label is drawn with a background color matching its trajectory color; if a label must be moved for readability, a thin leader line connects the label to the endpoint dot.

Fig.~\ref{fig:qualitative} illustrates the visualization.

\subsection{Prompt Design}
\label{sec:appendix_prompt_design}

\noindent \textbf{Separation of concerns.}
The system prompt enforces a safety-first policy and defines output format. The user prompt provides per-step state: the goal (if any), candidate count, and a table for displayed candidates including geometry (and optionally planner confidence, which we often hide to avoid anchoring).

\noindent \textbf{Short-horizon semantics.}
The prompt explicitly states that candidates cover only $\sim$4\,s and that the goal can be off-screen and far beyond the horizon; therefore the correct behavior is to pick a \emph{locally safe} candidate that makes progress, not to ``reach the goal'' in one step.

\subsection{Full System/User Prompt and Example Outputs}
\label{sec:appendix_full_prompt}

We include the full prompt in Fig.~\ref{fig:appendix_prompt_and_outputs}. The actual VLM request contains both the text below and the overlay image (and optionally a short stack of history frames).
The system prompt does not change across steps or scenarios, which allows us to leverage context caching offered by the model provider (e.g., Gemini's cached-content API) to improve throughput and reduce both cost and latency.

\subsection{Output Validation and Robust Parsing}
\label{sec:appendix_parsing}

To make execution and evaluation robust, we validate VLM outputs and normalize common formatting deviations.
The parser handles the following cases in order:
\begin{enumerate}[leftmargin=*,nosep]
    \item \emph{Code-fence stripping:} JSON wrapped in Markdown code fences (e.g., triple-backtick \texttt{json} blocks) is extracted before parsing.
    \item \emph{JSON object extraction:} the first \texttt{\{...\}} block is parsed; action-field values \texttt{select\_trajectory}, \texttt{select}, \texttt{stop}, and \texttt{halt} are all accepted.
    \item \emph{Bare integer fallback:} if the response is a single integer (no JSON), it is treated as a trajectory index.
    \item \emph{Index validation:} if the returned index is not in the set of displayed labels, the parser attempts a rank-based mapping (interpreting the integer as a 0-based row index into the candidate table). If mapping also fails, the output is treated as invalid.
\end{enumerate}
If parsing fails entirely or the index is out of range after mapping, we treat the step as invalid and fall back to a safe behavior (planner argmax or stop) in deployment.

\section{Closed-Loop Benchmark}
\label{sec:appendix_closed_loop}

\subsection{Simulator Overview}
\label{sec:appendix_closed_loop_sim}

We evaluate latency resilience in a lightweight closed-loop simulator that tracks a reference path extracted from real odometry. The simulator runs control at $\Delta t_\text{control}=\SI{0.1}{s}$, replanning at $\Delta t_\text{plan}=\SI{0.2}{s}$ (5\,Hz), with a 4\,s trajectory horizon (20 waypoints).

At each plan tick, we transform a fixed library of $K$ robot-frame candidate trajectories (we use a K-means medoid set; $K=12$ in our default experiments) into world frame using the current pose and select one candidate according to the evaluated policy; a tracking controller (pure pursuit) executes the chosen polyline under velocity and rate limits.

\noindent \textbf{Non-visual, kinematics-only design.}
Because the simulator's sole purpose is to study the effect of \emph{delay and fusion policies} on closed-loop control---not to evaluate perception---it is intentionally non-visual: there is no rendering, no physics engine, and no sensor simulation.
The robot state evolves under unicycle kinematics:
\begin{equation}
    \dot{x} = v\cos\theta, \quad
    \dot{y} = v\sin\theta, \quad
    \dot{\theta} = \omega,
\end{equation}
subject to velocity and rate limits ($v_{\max}=\SI{2.0}{m/s}$, $\omega_{\max}=\SI{0.5}{rad/s}$).
A pure-pursuit controller converts the selected polyline into $(v, \omega)$ commands at each control tick.
This lightweight design lets us sweep delay values, fusion parameters, and corruption levels over hundreds of episodes in minutes.

\subsection{Reference-Tracking Tasks}
\label{sec:appendix_closed_loop_tasks}

Each closed-loop task is built from a window of real odometry:
a reference polyline of duration $T_\text{ref}=\SI{18}{s}$ extracted from odometry, with
tasks sampled at stride $T_\text{stride}=\SI{5}{s}$.
We require each source episode to be at least \SI{18}{s} long and sample up to 1 task per episode (uniformly over the stride grid) from up to 100 episodes, yielding up to 100 reference-tracking tasks in total.

We primarily use an \emph{arclength} completion definition (progress along the reference polyline) with an independent simulation time budget (default 40\,s). This avoids coupling the evaluation to the original logging speed and focuses on path-following completion.

\subsection{Static Candidate Library}
\label{sec:appendix_closed_loop_candidates}

The candidate library contains $K=12$ robot-frame polylines, each resampled to 20 waypoints over a 4\,s horizon. We construct the library by clustering a large pool of logged trajectories and selecting representative medoids (k-means in trajectory space; one medoid per cluster).

\subsection{``Fake Planner'' (Corrupted Score Model)}
\label{sec:appendix_fake_planner}

We define an \textbf{oracle} candidate trajectory out of the candidate set as the one minimizing mean pointwise distance to the current reference segment. The \emph{planner score} is the same base objective plus controlled corruption to simulate mis-scoring:
\begin{itemize}[leftmargin=*,nosep]
    \item {Gaussian noise} with standard deviation $\sigma$ (default 1.0);
    \item {score temperature} $T_s$ (default 1.0; $<1$ sharper, $>1$ flatter), applied as division by $T_s$;
    \item {$\epsilon$-corruption}: with probability $\epsilon$ (default 0.3), the entire score vector is replaced with uniform random values, making the argmax effectively random for that tick.
\end{itemize}

\subsection{``Fake VLM'' (Delayed Oracle Model)}
\label{sec:appendix_fake_vlm}

We model the VLM as a delayed oracle: at request time $t_0$ it selects the oracle index for that reference segment, and the response is delivered at $t_0+\Delta t$ with fixed delay $\Delta t$. 

\subsection{Policies and Latency Handling}
\label{sec:appendix_closed_loop_policies}

We evaluate three families of policies:
(i)~\emph{direct execution} of stale VLM trajectories (\textit{VLM Hold} and \textit{VLM Stream});
(ii)~\emph{matching} the stale VLM trajectory to the closest current candidate (\textit{VLM Match});
and (iii)~\emph{fusion} policies that bias the planner selection toward the stale VLM intention while still choosing among current candidates (\textit{Score Fusion} / \textit{Probability Fusion}).

\noindent \textbf{Request scheduling and pipelining.}
We distinguish \emph{sequential} request policies (submit the next query only after receiving the previous response; single in-flight request) from \emph{streaming} request policies (submit at a fixed cadence; multiple pipelined in-flight requests). This separation isolates the effect of latency from throughput limitations.

\section{Real-World Deployment Setup}
\label{sec:appendix_real_world}

\subsection{System Architecture}
\label{sec:appendix_real_world_arch}

The real-world system follows a two-rate architecture: a fast onboard local planner continuously proposes short-horizon, dynamically feasible candidate trajectories, while a slower VLM is queried asynchronously to provide high-level intent in the form of trajectory selection (Sec.~\ref{sec:appendix_vlm_interface}).
Crucially, control and planning never block on the VLM response. Instead, the system (i) executes the planner in a receding-horizon loop and (ii) incorporates the most recent available VLM intent using the latency-handling policies described in the main paper (direct execution, matching, or fusion).

\noindent \textbf{Asynchronous execution and time alignment.}
Each VLM request is tagged with a monotonically increasing request ID and a timestamp corresponding to the camera frame used for the overlay. When the response arrives, the policy aligns it to the current planning tick using (a) the request ID and (b) the current candidate set, applying either:
(i) \emph{hold}-style execution (execute the stale intent directly when feasible),
(ii) \emph{match} (map the stale intent to the closest current candidate), or
(iii) \emph{fusion} (bias the current planner selection toward the stale VLM intent while still selecting among up-to-date candidates).
If no valid VLM output is available, the system falls back to a safe default (planner-only with conservative stopping).

\subsection{Robot, Sensors, and Control Rates}
\label{sec:appendix_real_world_hw}

We deploy on a wheeled delivery robot (differential-drive, four actuated wheels) with onboard odometry and a panoramic fisheye camera suite. The front fisheye camera serves as the primary perception input to both the overlay generation and the planner/VLM interface; GPS and odometry are used for state estimation and goal/route bookkeeping when needed.

\noindent \textbf{Perception streams.}
The system uses:
\begin{itemize}[leftmargin=*,nosep]
    \item front camera input at 5\,Hz (1920$\times$1080 raw stream);
    \item planner input resized to 512$\times$288 (deterministic fisheye$\rightarrow$pinhole conversion for planner input frames);
    \item overlay images rendered at a fixed resolution (e.g., 960$\times$540) consistent with Sec.~\ref{sec:appendix_overlay_design}.
\end{itemize}

\noindent \textbf{Planning and control loop.}
Candidate generation and waypoint updates run at 5\,Hz. The robot executes velocity commands derived from waypoints using a simple kinematic controller (differential-drive), with conservative limits:
\[
v_{\max}=\SI{0.5}{m/s}, \quad \omega_{\max}=\SI{0.5}{rad/s}.
\]
Odometry is consumed at its native robot rate to update the robot pose and to continuously update the tracking target during navigation.

\noindent \textbf{Pure pursuit tracking controller.}
The selected waypoint polyline is tracked by a pure pursuit controller.
The look-ahead distance is velocity-adaptive:
$L_d = \max(L_0,\; L_0 + k_v \cdot v)$,
with base look-ahead $L_0 = \SI{1.0}{m}$ and gain $k_v = 0.5$.
The curvature command is derived from the look-ahead point and converted to angular velocity $\omega$ for the differential-drive base.
The controller enforces hard limits on linear acceleration ($a_{\max}=\SI{1.0}{m/s^2}$), angular acceleration ($\dot{\omega}_{\max}=\SI{1.2}{rad/s^2}$), and lateral acceleration ($a_{\text{lat,max}}=\SI{0.8}{m/s^2}$) to ensure smooth, safe motion.

\subsection{VLM Query, Staleness Handling, and Safety Mechanisms}
\label{sec:appendix_real_world_vlm}

\noindent \textbf{Query scheduling.}
The VLM is queried asynchronously using the overlay image plus the text prompt (Sec.~\ref{sec:appendix_prompt_design}). We support two scheduling modes:
\begin{itemize}[leftmargin=*,nosep]
    \item \emph{Sequential} (used by \texttt{vlm\_hold} and \texttt{vlm\_hold\_match}): a single request is in flight at any time; the next request is submitted only after the previous response is received and processed. This maximizes freshness per response but limits throughput.
    \item \emph{Streaming} (used by \texttt{vlm\_stream}, \texttt{score\_fusion\_stream}, \texttt{prob\_fusion\_stream}): requests are submitted at a fixed cadence (default 1\,Hz) regardless of whether previous responses have arrived, allowing multiple in-flight requests. When responses return (potentially out of order due to variable network latency), the system adopts the \emph{newest} advice by query timestamp.
\end{itemize}

\noindent \textbf{Output validation.}
All VLM outputs are parsed and validated as in Sec.~\ref{sec:appendix_parsing}. Invalid outputs (non-integer index, out-of-range index, or unparsable formatting) are discarded and treated as missing.

\begin{figure}[H]
    \centering
    \includegraphics[width=\linewidth]{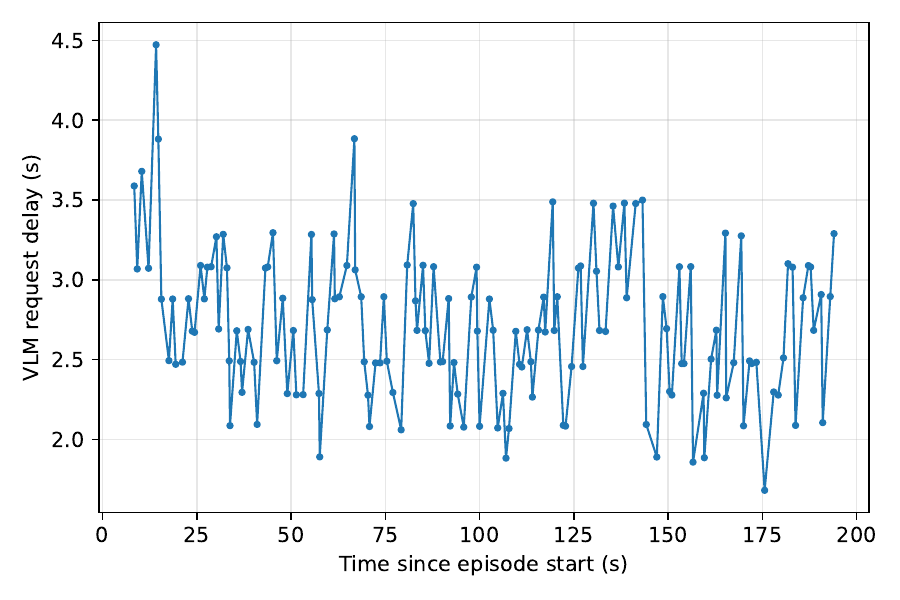}
    \caption{\textbf{VLM request--response latency during real-world deployment} (Probability Fusion + Streaming, Gemini 2.5 Flash Lite over 4G cellular). Each dot is one VLM response; the $y$-axis shows the wall-clock delay between request submission and response receipt. The median latency is $\approx$2.7\,s (min 1.7\,s, max 4.5\,s). The variation is driven by network jitter rather than model inference time. All responses arrive well within the \SI{5}{s} hard staleness timeout, confirming that VLM Streaming with fusion provides continuous guidance under realistic network conditions.}
    \label{fig:vlm_delay}
\end{figure}

\noindent \textbf{Hard staleness timeout and fail-safe.}
To prevent executing arbitrarily stale intent, we enforce a hard timeout: if the most recent valid VLM response is older than a fixed threshold ($\SI{5}{s}$) relative to the current control time, the system switches to a conservative behavior:
(i)~stop (preferred in crowded/uncertain scenes, used in ``VLM Hold'' and ``VLM Stream'' modes) or
(ii)~planner-only execution with conservative stopping enabled.
Fig.~\ref{fig:vlm_delay} shows the observed VLM request--response latency over a representative real-world deployment run; the majority of responses arrive within the \SI{5}{s} timeout window, validating our choice of threshold.

\noindent \textbf{Human-in-the-loop safety.}
All real-world runs include a trained safety operator with an immediate override capability (teleoperation or emergency stop). Any intervention immediately cancels the current VLM intent and returns control to a safe mode; the resulting takeover event is logged and used to compute the safety metrics reported in the main paper. Speed limits are always enforced, and the robot is operated only in pedestrian environments where a safe stop is feasible at all times.

\subsection{Evaluation Protocol and Metrics}
\label{sec:appendix_real_world_protocol}

\noindent \textbf{Environments and routes.}
Evaluation is conducted on outdoor pedestrian routes (e.g., sidewalks and campus pathways) containing natural obstacles such as pedestrians, curb cuts, surface boundaries (grass/planters), and intersections/forks. Each route is executed multiple times per method under similar conditions; all sensor streams, planner candidates, chosen indices, and operator interventions are logged with timestamps.

\noindent \textbf{Runs and completion.}
A run begins at a fixed start pose and continues until the robot reaches the route endpoint (within a small tolerance).
In our experimental protocol, every trial is completed regardless of the number of takeovers: when a takeover occurs, the safety operator manually guides the robot back to a safe pose and returns control to the autonomous policy. The run then continues from that point.
This ensures that all metrics (takeover rate, trajectory consistency, VLM--planner agreement) are computed over the \emph{full route} for every method, enabling fair comparison.

\section{Fusion Design Choices, Sensitivity, and Ablation}
\label{sec:appendix_sensitivity}

This appendix justifies the three design knobs of the fusion layer (Sec.~\ref{sec:fusion}) and reports the corresponding sensitivity sweeps and ablations.
All numbers are mean position error (m) from closed-loop simulation under the setup described in Sec.~\ref{sec:appendix_closed_loop}, averaged across 5 seeds and 3 reference-tracking tasks (forward, left-turn, right-turn) unless otherwise noted, with planner corruption $\epsilon{=}0.3$.

\subsection{Why Horizon-Truncated Similarity}
\label{sec:appendix_design_horizon}
A stale VLM trajectory of duration $H$ has, by the time the response arrives, already been consumed by $\Delta t$ seconds of robot motion, leaving an $H{-}\Delta t$ overlap with the fresh candidates.
We project the current robot pose onto the stale polyline, compute remaining arclength, and truncate the comparison to that overlap (Eq.~\ref{eq:similarity}).
Comparing the fresh candidate against the full stale polyline instead would waste overlap on a phantom future tail the VLM neither saw nor reasoned about.
Tab.~\ref{tab:sim_ablation} confirms this on the left-turn task at $\Delta t{=}2$\,s: horizon-aware similarity is the strongest setting for Score Fusion and is competitive for Probability Fusion under decay-off; the performance---Probability Fusion + horizon-aware + decay-on (3.565)---falls within the spread of the table and is dominated by the decay axis (Sec.~\ref{sec:appendix_design_decay}).

\subsection{Why Exponential Staleness Decay}
\label{sec:appendix_design_decay}
A single time constant $\tau_\text{decay}$ in $w(\Delta t)=\exp(-\Delta t/\tau_\text{decay})$ is more interpretable than a piecewise or polynomial schedule, and a soft cutoff matters because network latency is heavy-tailed (Fig.~\ref{fig:vlm_delay}): a hard threshold would flip advice between fully-trusted and ignored at the worst possible moments.
The right reading of decay is \emph{robustness insurance under real-world drift}, not an unconditional accuracy gain. In our delayed-oracle sim the VLM never changes its mind, so its advice is never wrong, and any time-discount over-penalises advice that is still useful---decay-off is therefore slightly better than decay-on at $\Delta t{=}2$\,s in Tab.~\ref{tab:sim_ablation}.
In real deployment, where a VLM's stale answer can become actively wrong as the scene evolves, decay is what bounds the damage.
The heatmap below sweeps $\tau_\text{decay}$ against the fusion weight $\lambda$; it is monotone in both axes, and the operating point used elsewhere in the paper sits inside the stable plateau rather than on a fragile peak.

\begin{table}[H]
\centering
\caption{Similarity-mode and staleness-decay ablation. Mean position error (m) on left-turn task, $\Delta t{=}2$\,s, averaged over 5 seeds, $\epsilon{=}0.3$. ``Decay on'' uses $\tau{=}3$\,s; ``decay off'' uses $\tau{\to}\infty$ (similarity term applied without time-discounting).}
\label{tab:sim_ablation}
\begin{tabular}{llcc}
\toprule
Policy & Similarity mode & Decay on ($\tau{=}3$) & Decay off ($\tau{\to}\infty$) \\
\midrule
Score Fusion & pointwise (arclen) & 3.521 & 3.530 \\
Score Fusion & candidate$\to$stale polyline & 3.505 & 3.501 \\
Score Fusion & \textbf{horizon-aware (ours)} & \textbf{3.485} & \textbf{3.456} \\
\midrule
Probability Fusion & pointwise (arclen) & 3.556 & 3.504 \\
Probability Fusion & candidate$\to$stale polyline & 3.530 & 3.470 \\
Probability Fusion & \textbf{horizon-aware (ours)} & \textbf{3.565} & \textbf{3.457} \\
\bottomrule
\end{tabular}
\end{table}

\subsection{Hyperparameter Heatmap: $\tau_\text{decay}$ vs.\ $\lambda$}
\label{sec:appendix_sensitivity_heatmap}

Tab.~\ref{tab:sim_heatmap} reports Score Fusion mean position error at $\Delta t{=}2$\,s, sweeping the staleness time constant $\tau_\text{decay}$ and the fusion weight $\lambda$.
The surface is monotone in both axes: error decreases as $\lambda$ increases (more VLM influence) and as $\tau_\text{decay}$ increases (slower staleness decay).
There is no pathological corner.
Our default operating point $(\tau{=}3,\lambda{=}1)$ sits inside the stable plateau rather than on a fragile peak, supporting the claim that the choice is not cherry-picked.

\begin{table}[H]
\centering
\caption{Score Fusion sensitivity to $\tau_\text{decay}$ and $\lambda$. Mean position error (m) at $\Delta t{=}2$\,s, averaged across 3 tasks $\times$ 5 seeds, $\epsilon{=}0.3$. Lower is better.}
\label{tab:sim_heatmap}
\begin{tabular}{c|cccccc}
\toprule
$\tau_\text{decay}$ \textbackslash\ $\lambda$ & 0.1 & 0.5 & 1 & 2 & 5 & 10 \\
\midrule
1\,s & 3.093 & 3.065 & 3.058 & 3.020 & 2.991 & 2.932 \\
2\,s & 3.070 & 3.036 & 3.012 & 2.984 & 2.896 & 2.781 \\
3\,s & 3.070 & 3.027 & 2.997 & 2.969 & 2.812 & 2.754 \\
5\,s & 3.065 & 3.016 & 2.990 & 2.932 & 2.792 & 2.720 \\
8\,s & 3.065 & 3.011 & 2.983 & 2.914 & 2.772 & \textbf{2.702} \\
\bottomrule
\end{tabular}
\end{table}

\section{Hyperparameters}
\label{sec:appendix_hparams}

\begin{table}[H]
\centering
\caption{Fusion hyperparameters (real-world deployment defaults).}
\label{tab:appendix_fusion_hparams}
\begin{tabular}{lcc}
\toprule
Parameter & Symbol & Value \\
\midrule
Staleness decay constant & $\tau$ & 5\,s \\
Similarity distance scale & $d_\text{scale}$ & 0.3\,m \\
Score Fusion weight & $\lambda$ & 1.0 \\
Probability Fusion weight & $\lambda$ & 3.0 \\
Similarity softmax temperature & $T_\text{vlm}$ & 1.0 \\
Hard staleness timeout & -- & 5\,s \\
\bottomrule
\end{tabular}
\end{table}

\section{Qualitative Results: When VLM Helps Most}
\label{sec:appendix_qualitative}

We observe the largest open-loop ADE improvements in scenarios where the planner proposes at least one good candidate but assigns it a low score due to limited semantic scene understanding. Common categories include:
\begin{itemize}[leftmargin=*,nosep]
    \item \textbf{Off-limits surface boundaries}: grass/dirt/planter edges where the planner argmax drifts off sidewalk but a safe sidewalk-centered candidate exists.
    \item \textbf{Pedestrian interactions}: yielding and choosing a path that maintains clearance (especially when candidates overlap in geometry but differ in safety).
    \item \textbf{Forks and junctions}: selecting the branch consistent with the goal direction while staying on the walkable surface.
    \item \textbf{Occlusion and ambiguity}: preferring conservative stop/slow candidates when visibility is poor.
\end{itemize}

Fig.~\ref{fig:quali_appendix} shows representative examples where the VLM selects a substantially better trajectory than the planner's top-scored candidate.
In the top-left panel, the planner drifts toward the edge of the sidewalk while the VLM centers the robot on the path (2.5\,m ADE improvement).
The top-right panel shows a similar surface-boundary scenario at night, where the VLM keeps the robot safely centered between the curb and wall (2.3\,m improvement).
The bottom-left panel illustrates a crosswalk approach: the VLM identifies a car and selects a trajectory that stays on the paved path toward the crosswalk (2.0\,m improvement).
In the bottom-right panel, the VLM maintains a respectful distance from a pedestrian on the sidewalk (1.1\,m improvement).
In all four cases, the planner's candidate set contains a good trajectory but the planner assigns it a low score; the VLM's semantic reasoning---identifying surface boundaries, vehicles, and pedestrians---recovers the gap.

\begin{figure*}[!t]
    \centering
    \includegraphics[width=\textwidth]{figs/quali_appendix.pdf}
    \caption{\textbf{Qualitative examples: when VLM selection helps most.}
    Four scenarios where the VLM outperforms the planner's argmax, with JSON outputs showing VLM reasoning.
    \textbf{Top-left}: sidewalk centering (VLM selects trajectory~38 over planner's~8; 2.5\,m ADE improvement).
    \textbf{Top-right}: night navigation near curb and wall (trajectory~32 vs.\ planner's~10; 2.3\,m improvement).
    \textbf{Bottom-left}: crosswalk approach with a car visible (trajectory~6 vs.\ planner's~10; 2.0\,m improvement).
    \textbf{Bottom-right}: pedestrian interaction on sidewalk (trajectory~41 vs.\ planner's~8; 1.1\,m improvement).
    In each case, the planner's candidate set contains the correct trajectory but the scoring function assigns it a low rank; the VLM's semantic understanding recovers the gap.}
    \label{fig:quali_appendix}
\end{figure*}

\clearpage
\onecolumn

\section{VLM Prompt}

\begin{center}
{\scriptsize
\captionof{figure}{\textbf{Trajectory-selection prompt (system/user text) and example outputs.} The actual request includes the overlay image in addition to this text.}
\label{fig:appendix_prompt_and_outputs}
}
\begin{scriptsize}
\begin{promptbox}
(*@\textbf{SYSTEM PROMPT}@*)
You are a navigation assistant controlling a ground robot.

Robot footprint: assume the robot is 0.80 m wide (use this as the clearance envelope). At each step, a local planner proposes multiple candidate trajectories (shown in an overlay image). Each candidate trajectory is a polyline of 20 waypoints sampled at 5 Hz (dt=0.2 s), representing ~4 seconds of future motion. Your job is to choose ONE action for the next step.

Task: select a candidate trajectory index that is SAFE and makes progress toward the goal. IMPORTANT: candidates are short-horizon (~4 seconds). You do NOT need to reach the goal in one step; instead choose a trajectory that moves toward the goal direction while staying safe.

Decision procedure:
1) Scene understanding first: identify sidewalk/path vs grass/dirt/planters, and any pedestrians/obstacles.
2) HARD REJECT any trajectory that goes onto grass/off-limits surface or too close to a pedestrian/obstacle.
3) Only among the remaining safe options, use goal geometry (progress/angle) as a tie-breaker.
4) If you are unsure whether a trajectory stays on sidewalk (ambiguous), choose a safer option or stop.

If none look safe, choose stop. If the image is missing/unclear or you cannot decide, choose stop. Sometimes a very short trajectory indicates a 'stop-like' action. In that case, prefer returning {"action":"select_trajectory","selected_index": <that short trajectory index>} rather than returning stop.

Common critical objects / hazards to consider (not exhaustive):
- people / pedestrians (including children)
- cars / trucks / buses / motorcycles
- bicycles / scooters
- intersections / crosswalks / merging traffic
- construction cones / barriers
- utility pole / signpost / bollard
- curb edge / drop-off / fall risk
- stairs / steep slope
- grass / off-limits area
- loose dirt / gravel / sand / heavy leaves (low traction / not a sidewalk)
- parked vehicles opening doors
- clutter / boxes / trash cans
- animals
- occlusion / blind corner / poor visibility

Intention / driving style:
- Do not drive/walk on grass.
- Prefer paved sidewalks/paths; avoid loose dirt, gravel, sand, or heavy leaf piles.
- Yield to humans and give them right-of-way.
- Stop at intersections when uncertain or when crossing traffic/pedestrians may be present.
- Avoid colliding with any object.
- If anything looks unsafe or ambiguous, prefer the safest option; if none look safe, choose stop.

Candidate table columns (planner confidence intentionally hidden):
- index: the trajectory label shown in the overlay image (use it for selected_index).
- traj_dist_m: total path length of the trajectory (meters).
- end_xy: trajectory endpoint in robot frame (x_forward, y_left) in meters.
- end_goal_dist_m: distance from trajectory endpoint to the goal (meters).
- goal_ang_diff_deg: absolute angle difference between endpoint direction and goal direction (degrees).
- progress_m: progress toward the goal = goal_distance - end_goal_dist_m (meters).
NOTE: end_goal_dist_m / goal_ang_diff_deg / progress_m are pure geometry and do NOT account for obstacles/off-limits areas.

Goal semantics:
- The magenta marker labeled 'GOAL' indicates the desired navigation target in the robot frame.
- The goal can be far beyond the 4-second candidate horizon and may be off-screen.
- Choose a candidate that moves TOWARD the goal (good progress/angle) while remaining safe.
- Do NOT treat the goal marker as a physical object you must reach immediately.
- IMPORTANT: do NOT choose a trajectory just because its index label is closest to the GOAL marker. Use the table + image.

Legend:
- Camera: front-facing fisheye (~100 deg FOV). Distortion near the image edges is normal.
- Colored polylines are candidate trajectories.
- Each trajectory endpoint is marked with a small colored dot.
- Each trajectory has a number label near its endpoint: that number is the candidate index.
- The number label background color matches the trajectory color (use this to match label and line).
- If a label is moved for readability, a thin leader line connects the label to the endpoint dot.
- The magenta marker labeled 'GOAL' is the goal (may be absent if no goal is provided).
- The robot is at the bottom of the image; forward is upward.

Return exactly ONE action:
Schema: {"critical_object":[...], "reason":"...", "action":"select_trajectory|stop", "selected_index":<int|null>}

Reason requirements:
- Keep "reason" short
- Use less than 50 words

Examples (copy this key order exactly):
{"critical_object": [], "reason": "Path 8 goes toward the goal and makes good progress while staying on sidewalk.", "action": "select_trajectory", "selected_index": 8}
{"critical_object": ["pedestrian"], "reason": "Pedestrian crossing ahead. Path 3 curves around them safely.", "action": "select_trajectory", "selected_index": 3}
{"critical_object": ["construction", "blocked_path"], "reason": "All trajectories lead into blocked area.", "action": "stop", "selected_index": null}

Rules:
- If action="select_trajectory", selected_index must be the trajectory index shown in the image.
- If action is "stop", set selected_index to null.
- If there are no critical objects, set "critical_object": [].
Return ONLY valid JSON. Do not include any extra text.

(*@\textbf{USER PROMPT (text portion; the request also includes the overlay image)}@*)
Goal:
- Text: <optional goal text>
- Vector (robot frame): x_forward=<gx> m, y_left=<gy> m
- Bearing: <deg> deg (0=forward, +left)
- Distance: <d_goal> m

Number of candidates shown: <K>

Candidate trajectories shown in image:
  index | traj_dist_m |      end_xy | end_goal_dist_m | goal_ang_diff_deg | progress_m
    ... |        ...m | ( ... , ... )|          ...m   |           ...deg  |     ...m

(*@\textbf{MODEL OUTPUT (examples)}@*)
{"critical_object": [], "reason": "Stays on sidewalk and makes progress toward the goal.", "action": "select_trajectory", "selected_index": 11}
{"critical_object": ["pedestrian"], "reason": "Pedestrian ahead; no candidate maintains safe clearance.", "action": "stop", "selected_index": null}
\end{promptbox}
\end{scriptsize}
\end{center}
\twocolumn
\clearpage

\end{document}